\documentclass[runningheads]{llncs}
\usepackage{graphicx}
\usepackage{amsmath,amssymb} % define this before the line numbering.

\usepackage{color}
\usepackage[width=122mm,left=12mm,paperwidth=146mm,height=193mm,top=12mm,paperheight=217mm]{geometry}

\usepackage{cite}
\usepackage[colorlinks,citecolor={green}]{hyperref}  
\usepackage{bm}

\usepackage{algorithm,algcompatible,amssymb,amsmath}
\usepackage{multirow}
\usepackage{booktabs}
\usepackage{wrapfig}
\usepackage{sidecap}

\algnewcommand\algorithmicto{\textbf{to}}
\algnewcommand\RETURN{\State \textbf{return} }

\algnewcommand\algorithmicinput{\textbf{Input:}}
\algnewcommand\INPUT{\item[\algorithmicinput]}

\algnewcommand\algorithmicoutput{\textbf{Output:}}
\algnewcommand\OUTPUT{\item[\algorithmicoutput]}

\algnewcommand\algorithmicinitialize{\textbf{Initialize:}}
\algnewcommand\INITIALIZE{\item[\algorithmicinitialize]}

\definecolor{darkgreen}{rgb}{0.0, 0.5, 0.0}
\newcommand{\eg}{\emph{e.g. }}
\newcommand{\ie}{\emph{i.e. }}
\newcommand{\etal}{\emph{et al.}}

\begin{document}
% \linenumbers
\pagestyle{headings}
\mainmatter

\title{TS\textsuperscript{2}C: Tight Box Mining with Surrounding Segmentation Context for Weakly Supervised Object Detection} % Replace with your title

\titlerunning{TS\textsuperscript{2}C}
% Replace with a meaningful short version of your title

\authorrunning{Y. Wei et al}
% Replace with shorter version of the author list. If there are more authors than fits a line, please use A. Author et al.

\author{Yunchao Wei$^{1}$ \and Zhiqiang Shen$^{1, 2 *}$ \and Bowen Cheng$^{1 *}$ \thanks{* denotes equal contribution} \and Honghui Shi$^{3}$ \and \\
	 Jinjun Xiong$^{3}$ \and Jiashi Feng$^{4}$ \and Thomas Huang$^{1}$}

%Please write out author names in full in the paper, i.e. full given and family names. 
%If any authors have names that can be parsed into FirstName LastName in multiple ways, please include the correct parsing, in a comment to the volume editors:
%\index{Lastnames, Firstnames}
%(Do not uncomment it, because you may introduce extra index items if you do that, we will use scripts for introducing index entries...)

\institute{
	{\small $^{1}$University of Illinois at Urbana–Champaign, IL, USA} \\
	\email{ \{yunchao, shen54, bcheng9, t-huang1\}@illinois.edu} \\
	{\small $^{2}$Fudan University, Shanghai, China} \\
	{\small $^{3}$IBM T.J. Watson Research Center, Yorktown Heights, USA} \\
	\email{Honghui.Shi@ibm.com} \email{jinjun@us.ibm.com} \\
	{\small $^{4}$National University of Singapore, Singapore, Singapore} \\
	\email{elefjia@nus.edu.sg}
}

\renewcommand\footnotemark{}
\renewcommand\footnoterule{}
\maketitle

\begin{abstract}
	This work provides a simple approach to discover \emph{tight} object bounding boxes with only image-level supervision, called \textbf{T}ight box mining with \textbf{S}urrounding \textbf{S}egmentation \textbf{C}ontext (TS\textsuperscript{2}C). We observe that object candidates mined through current multiple instance learning methods are usually trapped to discriminative object parts, rather than the entire object. TS\textsuperscript{2}C leverages  surrounding segmentation context derived from weakly-supervised segmentation to suppress such low-quality distracting candidates and boost the high-quality ones. Specifically, TS\textsuperscript{2}C is developed based on two key properties of desirable bounding boxes: 1) high purity, meaning most pixels in the box are with high object response, and 2) high completeness, meaning the box covers high object response pixels comprehensively. With such novel and computable criteria, more tight candidates can be discovered for learning a better object detector. With TS\textsuperscript{2}C, we obtain 48.0\% and 44.4\% mAP scores on VOC 2007 and 2012 benchmarks, which are the new state-of-the-arts.
\end{abstract}

%%%%%%%%% BODY TEXT
%-------------------------------------------------------------------------
\section{Introduction}
\begin{figure}
	\centering
	\includegraphics[width=0.65\textwidth]{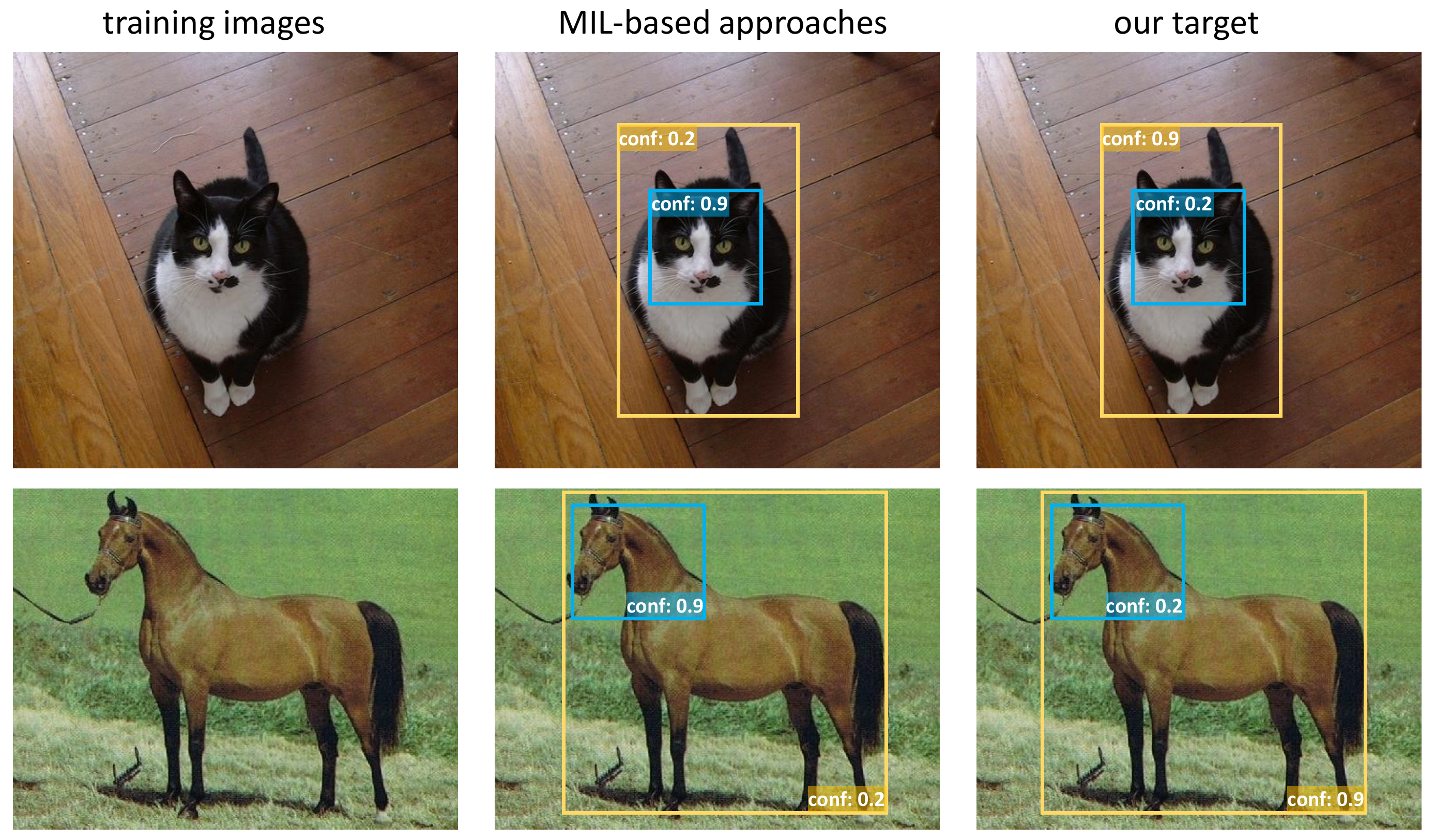}
	\caption{Comparison of MIL-based approaches and our target. MIL-based approaches tend to assign high confidence to discriminative parts (blue boxes) of target objects. Our target is to alleviate such cases and lift the confidence of the tight ones (yellow boxes). Best viewed in color.}
	\label{fig:motivation}
%    \vspace{-4mm}
\end{figure}

Weakly Supervised Object Detection (WSOD)~\cite{wang2014weakly,cinbis2017weakly,li2016weakly,jie2017deep,bilen2016weakly,kantorov2016contextlocnet,tang2017multiple,diba2016weakly,lai2017saliency,teh2016attention,zhang2018adversarial,zhang2018self,liang2015towards,zhao2018weakly} aims to detect objects only using image-level annotations for supervision. Despite  remarkable progress, existing approaches still have difficulties in accurately identifying tight boxes of target objects with only image-level annotations, thus their performance is inferior to the fully supervised counterparts~\cite{girshick2015fast,ren2015faster,liu2016ssd,redmon2016yolo9000,shen2017DSOD,cheng2018revisiting,li2017attentive}.

To localize objects with weak supervision information, one popular solution is to apply Multiple Instance Learning (MIL) for mining high-confidence region proposals~\cite{zitnick2014edge,uijlings2013selective} with positive image-level annotations. However, MIL usually discovers the most discriminative part of the target object (\eg the head of a cat)  rather than the entire object region, as shown in Figure~\ref{fig:motivation}. This inability of providing the complete object severely limits its effectiveness for WSOD. To address this issue, Li \etal~\cite{li2016weakly} exploited the contrastive relationship between a selected region and its mask-out image for proposal selection. Nevertheless, the mask-out strategy fails for multi-instance cases. The selector is easily confused by remained instances with high responses, even though the correct object has been masked out. 

\begin{figure*}[t]
	\centering
	\includegraphics[width=1\textwidth]{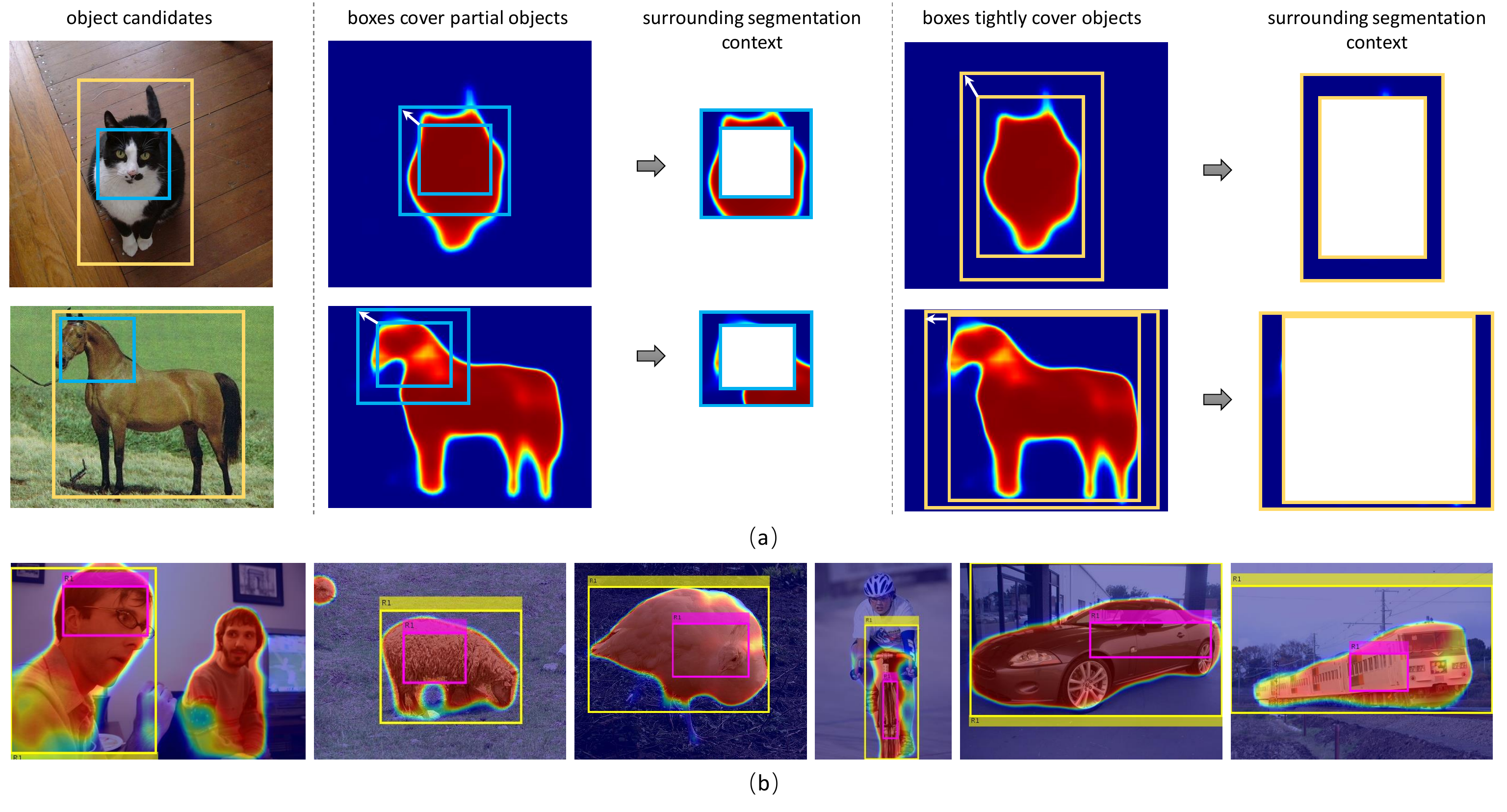}
	\caption{(a) Motivation of the proposed TS\textsuperscript{2}C: fewer high response pixels on the segmentation confidence map are included by enlarging higher-quality boxes of object candidates (the yellow one) compared with partial bounding boxes (the blue one). (b) Comparison of the rank 1 proposal using the strategy proposed by~\cite{diba2016weakly} (magenta boxes) and ours (yellow boxes). Best viewed in color.}
	\label{fig:comp}
%	\vspace{-3mm}
\end{figure*}

Recently, some weakly supervised semantic segmentation approaches~\cite{kolesnikov2016seed,wei2017object,wei2018revisiting,wei2015stc} have demonstrated promising performance. Utilizing the inferred segmentation confidence maps, Diba \etal~\cite{diba2016weakly} presented a cascaded approach that leverages segmentation knowledge to filter noisy proposals and achieves competitive detection results. However, we argue that their solution is sub-optimal and insufficient as it only considers the segmentation confidence \emph{inside} the proposal boxes, thus is unable to  filter high-response fragments of object parts, as the {{magenta}} boxes shown in Figure~\ref{fig:comp} (b). 

In this work, we propose a principled and more effective approach, compared with \cite{diba2016weakly}, to mine tight object boxes by exploiting segmentation confidence maps in a creative way, aiming for addressing the challenging WSOD problems. Our approach is motivated by the following observations, as illustrated by  two examples  in Figure~\ref{fig:comp} (a). We use {{blue}} and {{yellow}} to encode two kinds of boxes, which partially and tightly cover objects respectively. Based on the semantic segmentation confidence maps obtained in a weakly supervised manner,  many pixels surrounding the {{blue}} boxes have high predicted segmentation confidence, while very few high-confidence pixels are included in the surrounding context for the {{yellow}} ones  of higher tightness. We find that a desirable tight object box generally needs to satisfy two properties based on segmentation context:
\begin{itemize}
	\item \emph{Purity}: most pixels inside the box should have high confidence scores, which guarantees that the box is located around the target object;
	\item \emph{Completeness}: very few pixels are with high confidence scores in the surrounding context of the target box.
\end{itemize}

Based on these properties, we devise a simple yet effective approach, named Tight box mining with Surrounding Segmentation Context (TS\textsuperscript{2}C), to efficiently select object candidates of high quality from thousands of candidates. Specifically, the proposed TS\textsuperscript{2}C examines two kinds of regions for evaluating the tightness of bounding boxes: 1) the region included in the box and 2) the region surrounding the box. It computes objectness scores of the two regions by averaging the corresponding pixel confidence values on the segmentation maps. Tight boxes are expected to be with high and low objectness values of the two kinds of regions simultaneously. Thus, the difference of two objectness scores is then taken as the quality metric on the final tightness for ranking object candidates. Figure~\ref{fig:comp} (b) shows the top 1 object candidate inferred by the proposed TS\textsuperscript{2}C. We can see that our approach is more effective for mining tight object boxes than~\cite{diba2016weakly}. Moreover, our proposed TS\textsuperscript{2}C is generic and can be easily integrated into any WSOD framework by introducing a parallel semantic segmentation branch for class-specific confidence map prediction. 
%In addition, it is a common operation to leverage the pseudo object bounding boxes mined by WSOD approaches~\cite{jie2017deep,tang2017multiple} as ground-truth to learn a more powerful Fast-RCNN detector. Beyond the proposed TS\textsuperscript{2}C, we also present a step-wise training strategy to relieve the risk cased by false positive pseudo object boxes during learning the Fast-RCNN detector.
Benefiting from our TS\textsuperscript{2}C, we achieve 48.0\% and 44.4\% mAP scores on the challenging Pascal VOC 2007 and VOC 2012 benchmarks, which are the new state-of-the-arts in the WSOD community.

%-------------------------------------------------------------------------
\section{Related Work}
Multiple Instance Learning (MIL) provides a suitable way for formulating and solving WSOD. In specific, if an image is annotated with a specific class, at least one proposal instance from the image is positive for this class; and no proposal instance is positive for unlabeled classes. Previous works on applying MIL to WSOD can be roughly categorized into two-step~\cite{wang2014weakly,cinbis2017weakly,li2016weakly,jie2017deep} and end-to-end ~\cite{bilen2016weakly,kantorov2016contextlocnet,tang2017multiple,diba2016weakly,lai2017saliency,teh2016attention} based approaches.
%\noindent

\textbf{Two-step approaches} first extract proposal representation leveraging hand-crafted features or pre-trained CNN models and employ MIL to select the best object candidate for learning the object detector. For instance, Wang \etal~\cite{wang2014weakly} presented a latent semantic clustering approach to select the most discriminative cluster for each category. Cibis \etal~\cite{cinbis2017weakly} learned a multi-fold MIL detector by re-labeling proposals and re-training the object classifier iteratively. Li \etal~\cite{li2016weakly} first trained a multi-label classification network on entire images and then selected class-specific proposal candidates using a mask-out strategy, followed by MIL for learning a Fast R-CNN detector. Recently, Jie \etal~\cite{jie2017deep} took a similar strategy as Li \etal~\cite{li2016weakly} and proposed a more robust self-taught approach to learn a detector by harvesting more accurate supportive proposals in an online manner. However, splitting the WSOD into two steps results in a non-convex optimization problem, making such approaches trapped in local optima.

\textbf{End-to-end approaches} combine CNNs and MIL into a unified framework for addressing WSOD. Oquab \etal~\cite{oquab2015object} and Wei \etal~\cite{wei2015hcp} adopted a similar strategy to learn a multi-label classification network with max-pooling MIL. The learned classification model was then applied to coarse object localization~\cite{oquab2015object}. Bilen \etal~\cite{bilen2016weakly} proposed a novel Weakly Supervised Deep Detection Network (WSDDN) including two key streams, one for classification and the other for object localization. The outputs of these two streams are then combined for better rating the objectness of proposals. Based on WSDDN, Kantorov \etal~\cite{kantorov2016contextlocnet} proposed to learn a context-aware CNN with contrast-based contextual modeling. Both~\cite{kantorov2016contextlocnet} and our approach employ proposal context to identify high-quality proposals. However, \cite{kantorov2016contextlocnet} exploits inside/outside context features of each bounding box for learning to classification, in contrast, we leverage objectness scores obtained by segmentation confidence maps to pick out tight candidates. Recently, Tang \etal~\cite{tang2017multiple} also employed WSDDN as the basic network and augmented it with several Online Instance Classifier Refinement (OICR) branches, which is the state-of-the-art on the challenging WSOD task. In this work, we employ both WSDDN and OICR to develop our framework where the proposed TS\textsuperscript{2}C is leveraged to further improve performance. Both~\cite{diba2016weakly} and our approach utilizes object segmentation knowledge to benefit WSOD. However, Diba \etal~\cite{diba2016weakly} only considered the confidence of pixels included in the bounding box for rating the proposal objectness, which is not as effective as ours.

Beyond the above mentioned related works, some fully-supervised object detection approaches~\cite{gidaris2015object,chen20153d,zhu2015segdeepm,li2017attentive} also exploit contextual information of bounding boxes for benefiting object detection. Both Chen \etal~\cite{chen20153d} and Li \etal~\cite{li2017attentive} leveraged information of enlarged contextual proposals to enhance the accuracy of the classifier. Zhu \etal~\cite{zhu2015segdeepm} proposed to use a pool of segments obtained in the bottom-up manner to obtain better detection boxes. Our TS\textsuperscript{2}C is totally different from these works in terms of both motivation and methodology. In particular, our motivation is to employ surrounding segmentation context to suppress these false positive objects parts. In addition, our approach can be easily embedded into any WSOD framework to make a further performance improvement.
\begin{figure*}[t]
	\centering
	\includegraphics[width=0.9\textwidth]{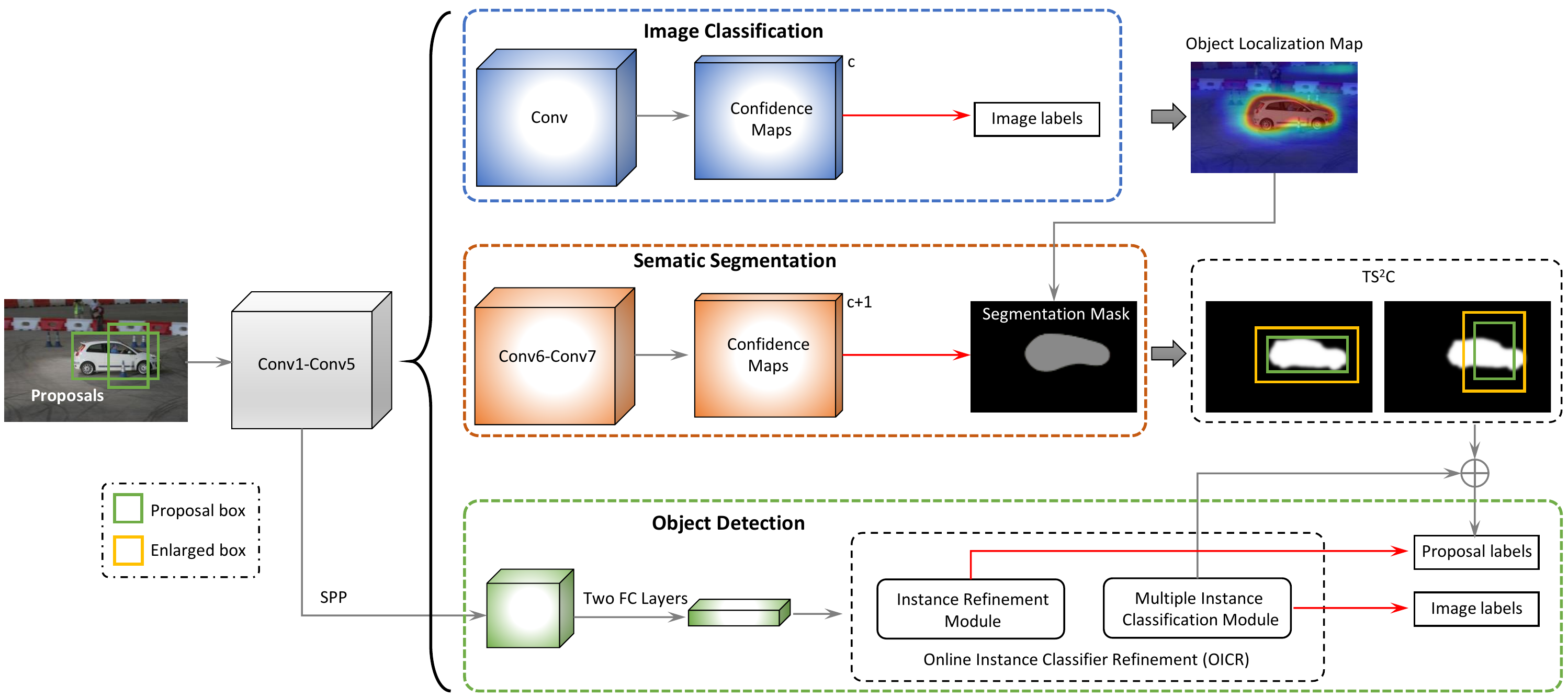}
	\caption{Overview of the proposed TS\textsuperscript{2}C for weakly supervised object detection. Several convolutional layers are leveraged to extract the intermediate features of an input image. The entire feature maps are firstly fed into a \emph{Classification} branch to produce object localization maps corresponding to image-level labels. We then employ the localization maps to generate the segmentation masks, which serve as supervision to learn the \emph{Segmentation} branch. Based on the segmentation confidence maps, we utilize TS\textsuperscript{2}C to evaluate the objectness scores of proposals according to their purity and completeness, which collaborates with the OICR~\cite{tang2017multiple} for training the \emph{Detection} branch.}
	\label{fig:framework}
%	\vspace{-3mm}
\end{figure*}
%-------------------------------------------------------------------------
\section{The Proposed Approach}
We show the overall architecture of the proposed approach in Figure~\ref{fig:framework}. It  consists of three key branches, \ie image classification,  semantic segmentation and  object detection. In particular, the \emph{Classification} branch is employed to generate class-specific localization maps. Following the previous weakly supervised semantic segmentation approaches~\cite{wei2016learning}, we leverage the inferred localization maps to produce pseudo segmentation masks of training images, which are then used as supervision to train the \emph{Segmentation} branch. The segmentation confidence maps from the \emph{Segmentation} branch are then employed to evaluate  objectness scores of the proposals according to the proposed TS\textsuperscript{2}C, which finally collaborates with the \emph{Detection} branch for learning an improved object detector. The overall framework is trained by minimizing the following composite loss functions from the three branches using stochastic gradient descent:
\begin{equation}
%L = L_{CLS} + L_{SEG} + L_{DET}.
L = L_{cls} + L_{seg} + L_{det}.
\end{equation}
We will introduce each branch below and then elaborate on details of  TS\textsuperscript{2}C.
\subsection{Classification for Object Localization}
\label{sec:cls}
Inspired by~\cite{zhou2015cnnlocalization,diba2016weakly,lin2013network}, the fully convolutional network along with the Global Average Pooling (GAP) operation is able to generate class-specific activation maps, which can provide coarse object localization prior. We conduct experiments on Pascal VOC benchmarks, in which each training image is annotated with one or several labels. We thus treat the classification task as a separate binary classification problem for each class. Following~\cite{oquab2015object}, the loss function $L_{cls}$ is thus defined as a sum of $C$ binary logistic regression losses.

\subsection{Weakly Supervised Semantic Segmentation}
The \emph{Classification} branch can produce localization cues for foreground objects. We assign the pixels with values on the class-specific confidence map larger than a pre-defined normalized threshold (\ie $\geq$0.78) with the corresponding class label. Beyond the object regions, background localization cues are also needed for training the segmentation branch. Motivated by~\cite{kolesnikov2016seed,wei2015stc,wei2017object,wei2018revisiting}, we leverage the saliency detection technology~\cite{xiao2018deep} to produce the saliency map for each training image. Based on the generated saliency map, we choose the pixels with low normalized saliency values (\ie $\leq$0.06) as background. However, both the class-specific confidence map and the saliency map are not accurate enough to guarantee a high-quality segmentation mask. To alleviate the negative effect caused by falsely assigned pixels, we ignore the ambiguous pixels during training the \emph{Segmentation} branch, including 1) pixels that are not assigned semantic labels, 2) foreground pixels of different categories that are in conflict, and 3) low-saliency pixels that fall in the foreground pixels. With the produced pseudo segmentation mask, we train the \emph{Segmentation} branch with pixel-wise cross-entropy loss $L_{seg}$, which is widely adopted by fully-supervised schemes~\cite{2015-long,chen2014semantic}.

\subsection{Learning Object Detection with TS\textsuperscript{2}C }
\label{sec:ts2c}
For each training or test image, Selective Search~\cite{uijlings2013selective} is employed to generate object proposals and Spatial Pyramid Pooling (SPP)~\cite{he2014spatial} is leveraged to generate constant size feature maps for different proposals. Our TS\textsuperscript{2}C aims to select high-quality object candidates from thousands of candidates to improve the effectiveness of training, which can be easily implanted into any WSOD framework. We choose the state-of-the-art Online Instance Classifier Refinement (OICR)~\cite{tang2017multiple} as the backbone of the \emph{Detection} branch, which collaborates with the proposed TS\textsuperscript{2}C for learning a better object detector. In the following, we will first make a brief introduction of OICR, and then explain how to leverage our TS\textsuperscript{2}C to benefit the learning process of WSOD.
%\begin{figure}[t]
%	\centering
%	\includegraphics[width=0.6\textwidth]{fig2/details.pdf}
%	\caption{Details of (a) Multiple Instance Classification Module and (b) Instance Refinement Module in TS\textsuperscript{2}C.}
%	\label{fig:details}
%\end{figure}

\subsubsection{OICR}
\begin{wrapfigure} {l} {0.5\textwidth}
%	\vspace{-3mm}
	\includegraphics[width=0.49\textwidth]{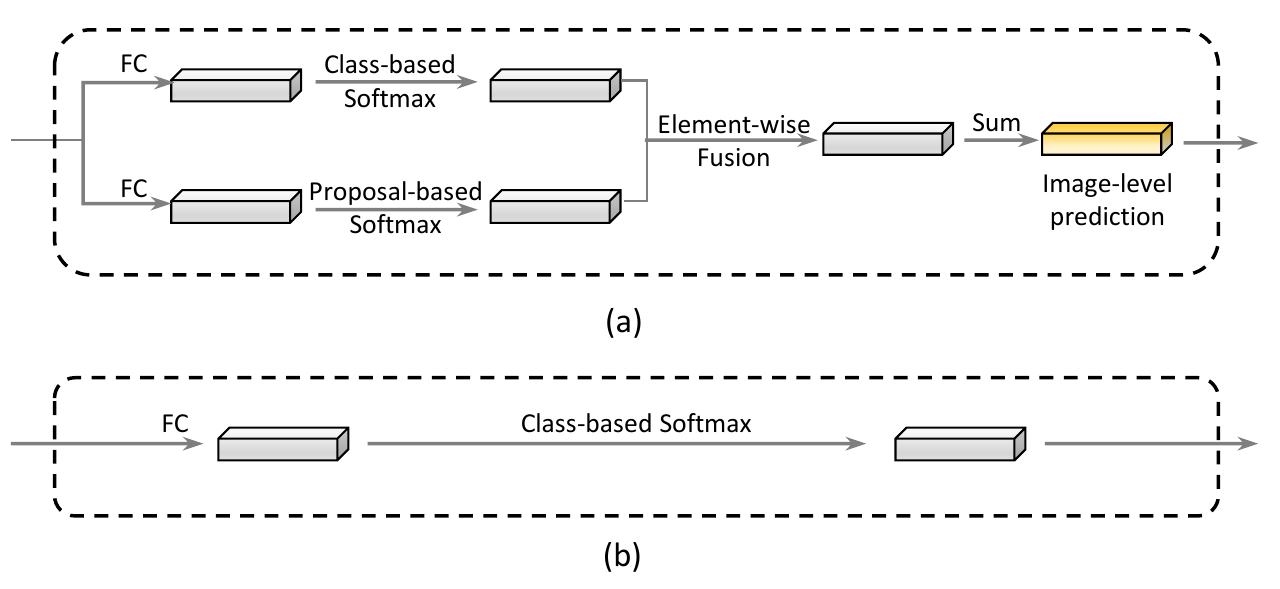}
	\caption{Details of (a) Multiple Instance Classification Module and (b) Instance Refinement Module in TS\textsuperscript{2}C.}
	\label{fig:details}
%	\vspace{-3mm}
\end{wrapfigure}

As shown in Figure~\ref{fig:framework}, the OICR mainly includes two modules, \ie multiple instance classification and instance refinement. In particular, the multiple instance classification module is inspired from~\cite{bilen2016weakly}, which includes two branches to extract parallel data streams from the input features pooled by SPP, as shown in Figure~\ref{fig:details} (a). The upper stream conducts softmax operation on each individual proposal for classification. The bottom stream estimates a probability distribution over all candidate proposals using softmax, which indicates the contribution of each proposal to classifier decision for each class. Therefore, these two streams provide classification-based and localization-based features for each proposal. Both inferred scores are then fused with element-wise product operation and finally aggregated into image-level prediction by sum-pooling over all proposals. With the supervision of image-level annotations, the multiple instance classification module can be learned with binary logistic regression losses as detailed in Section~\ref{sec:cls}. 

By leveraging multiple instance classification module as a basic classifier for obtaining initial classification scores for each proposal, progressive refinement is then conducted via the instance refinement module, as detailed in Figure~\ref{fig:details} (b). In particular, the instance refinement module first selects the top-scoring proposal of each image-level label. Those proposals with high spatial overlap scores over the top-scoring one are then labeled correspondingly. The idea behind such a module is that the top-scoring proposal may only contain part of a target object and its adjacent proposals may cover more object regions. Benefiting from both two modules embedded in the OICR, each proposal is assigned with a pseudo class label, which is then employed as supervision for learning detection with the softmax cross-entropy loss~\cite{girshick2014rich,girshick2015fast,ren2015faster}. To address the initialization issue (\ie  the classifier cannot well recognize proposals with randomly initialized parameters at the beginning of training), OICR adopts a weighted loss by assigning different weights to different proposals during different training iterations. Thus, the $L_{det}$ is composed of binary logistic regression losses for image-level classification and softmax cross-entropy loss for proposal-level classification. Please refer to~\cite{tang2017multiple} for more details. 

\subsubsection{Problems}
However, such progressive refinement operation of OICR highly relies on the quality of initial object candidates from the multiple instance classification module. This means without reasonable object candidates received from the multiple instance classification module for initialization, the following progressive refinement strategy of OICR cannot find the correct proposals with high IoU scores over ground-truth bounding boxes. This brings a critical risk: if the multiple instance classification module fails to produce reasonable object candidates then the OICR cannot recall the missed object with any hope. We propose to reduce such a risk by designing an objectness rating approach from a totally new perspective. In particular, we detail our proposed TS\textsuperscript{2}C that rates the proposals' objectness from the segmentation view in the following. 

\begin{figure}[t]
	\centering
	\includegraphics[width=0.8\textwidth]{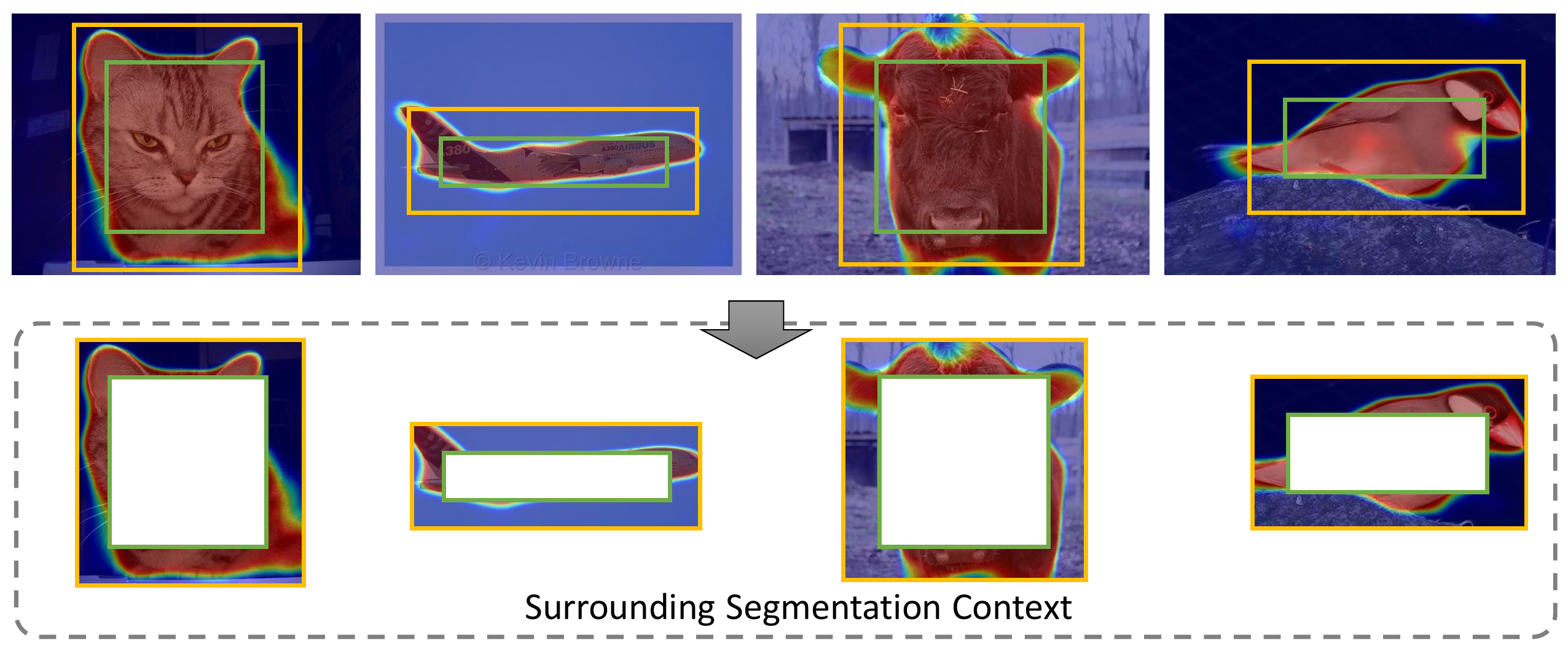}
	\caption{Motivation of the conditional average strategy: only a small number of pixels belong to objects in the surrounding regions. To promote the objectness score of surrounding context, we only employ pixels with large confidence values (highlighted by red color) for conducting average calculation. Best viewed in color.}
	\label{fig:condavg}
%    \vspace{-3mm}
\end{figure}

\subsubsection{TS\textsuperscript{2}C for Learning Detection}
As shown in Figure~\ref{fig:framework}, TS\textsuperscript{2}C uses the segmentation confidence maps from the \emph{Segmentation} branch to rate the proposal objectness. We consider $x_i (i = 1 \cdots n)$ as one proposal from a given training image annotated by class $c$. Let $H_c$ denote the confidence map of category $c$ predicted by the semantic \emph{Segmentation} branch. For $x_i$, we calculate objectness scores of both the region inside the box $P_I$ and the surrounding context $P_S$ between $x_i$ and the corresponding enlarged one. Let $avg(H_c, x_i)$ denote the operation of computing $P_I$, which takes all pixel values included in $x_i$ into account. $P_I$ of a large value can guarantee that $x_i$ is around the target object. To obtain a robust surrounding objectness score $P_S$, we adopt a conditional average strategy $\hat{avg}(H_c, x_i)$. As shown in Figure~\ref{fig:condavg}, many surrounding regions of negative candidates include a large number of un-related (\ie background) pixels, which are with low confidence scores. Therefore, the resulted objectness score will be small if we average all the pixel values for computing $P_S$, in a similar way as for $P_I$. However, we expect the value of $P_S$ to be large, so that negative candidates of such cases can be suppressed by $P_I-P_S$. To this end, we first rank the pixels in the surrounding region according to their confidence scores and the conditional average strategy only employs the first 50\% pixels for calculating the objectness score. Then, the objectness score $O(x_i)$ of the proposed TS\textsuperscript{2}C is finally calculated as
\[
\begin{aligned}
O(x_i) = P_I - P_S = avg(H_c, x_i) - \hat{avg}(H_c, x_i).     
\end{aligned}
\]
We rank all the object candidates according to $O(x_i)$ and build a candidate pool by selecting the top two hundred proposals, collaborating with the OICR for learning a better detector. As shown in Figure~\ref{fig:framework}, $\oplus$ means the OICR will only select object candidates from the pool produced by TS\textsuperscript{2}C for the following training process. 

During the testing stage, we ignore the \emph{Classification} and \emph{Segmentation} branches, and leverage the classification outputs from the instance refinement module to obtain the final detection results.

%-------------------------------------------------------------------------
\section{Experiments}
\subsection{Datasets and Evaluation Metrics}
\noindent \textbf{Datasets} We conduct experiments on Pascal VOC 2007 and 2012 datasets~\cite{2010-pascal}, which are the two most widely used benchmarks for weakly supervised object detection. For VOC 2007, we train the model on the \emph{trainval} set (5,011 images) and evaluate on the \emph{test} set (4,096 images). We also make extensive ablation analysis on VOC 2007 to verify the effectiveness of some settings. For VOC 2012, we train the model on the \emph{trainval} set (11,540 images) and evaluate on \emph{test} set (10,991 images) by submitting the testing result to the evaluation server.

\noindent \textbf{Metrics} Following~\cite{jie2017deep,tang2017multiple,diba2016weakly}, we adopt two metrics for evaluation, \ie mean average precision (mAP) and correct localization (CorLoc)~\cite{deselaers2012weakly}, for evaluation on \emph{test} and \emph{trainval} sets, respectively. Both two metrics employ the same threshold of bounding box overlaps with ground-truth boxes, \ie IoU $>=$ 0.5.

\subsection{Implementation Details}
We use the object proposals generated by Selective Search~\cite{uijlings2013selective}, and adopt the VGG16 network~\cite{simonyan2014very} pre-trained on ImageNet~\cite{2009-imagenet} as the backbone of the proposed framework. We employ the Deeplab-CRF-LargeFOV~\cite{chen2014semantic} model to initialize the corresponding layers in the segmentation branch. For the newly added layers, the parameters are randomly initialized with a Gaussian distribution $\mathcal{N}(\mu, \delta)(\mu=0, \delta=0.01)$. We take a mini-batch size of 2 images and set the learning rates of the first 40K and the following 30K iterations as 0.001 and 0.0001 respectively. During training, we take five image scales $\{480, 576, 688, 864, 1200\}$ for data augmentation. For TS\textsuperscript{2}C, we adopt an enlarged ratio of 1.2 to obtain the surrounding context, which is further employed for evaluating completeness of object candidates. Our experiments use the OICR~\cite{tang2017multiple} code, which is implemented based on the publicly available Caffe~\cite{jia2014caffe} deep learning framework. All of our experiments are run on NVIDIA TITAN X PASCAL GPUs.

%%%%%%%%% BODY TEXT

\begin{table*}[tb]\setlength{\tabcolsep}{2.4pt}
	\centering
	\caption{Comparison of detection average precision (AP) (\%) on PASCAL VOC.}
	\label{tab:det}
	\resizebox{1\textwidth}{!}{
		\begin{tabular}{lcccccccccccccccccccc|c}
			\toprule
			Method & \rotatebox{90}{plane} &   \rotatebox{90}{bike} &   \rotatebox{90}{bird} &   \rotatebox{90}{boat} &   \rotatebox{90}{bottle} &   \rotatebox{90}{bus} &   \rotatebox{90}{car} &   \rotatebox{90}{cat} &   \rotatebox{90}{chair} &   \rotatebox{90}{cow} &   \rotatebox{90}{table} &   \rotatebox{90}{dog} &   \rotatebox{90}{horse} &   \rotatebox{90}{motor} &   \rotatebox{90}{person} &   \rotatebox{90}{plant} &   \rotatebox{90}{sheep} &   \rotatebox{90}{sofa} &   \rotatebox{90}{train} &   \rotatebox{90}{tv} &  \rotatebox{90}{mAP}  \\
			\midrule
			\multicolumn{22}{l}{Comparisons on VOC 2007:}  \\
			Bilen \cite{bilen2014weakly} & 42.2 &43.9 &23.1 &9.2 &12.5 &44.9 &45.1 &24.9 &8.3 &24.0 &13.9 &18.6 &31.6 &43.6 &7.6 &20.9 &26.6 &20.6 &35.9 &29.6 &26.4\\
			
			Bilen \cite{bilen2015weakly} & 46.2 &46.9 &24.1 &16.4 &12.2 &42.2 &47.1 &35.2 &7.8 &28.3 &12.7 &21.5 &30.1 &42.4 &7.8 &20.0 &26.8 &20.8 &35.8 &29.6 &27.7\\
			
			Cinbis \cite{cinbis2017weakly} & 39.3 &43.0 &28.8 &20.4 &8.0 &45.5 &47.9 &22.1 &8.4 &33.5 &23.6 &29.2 &38.5 &47.9 &20.3 &20.0 &35.8 &30.8 &41.0 &20.1 &30.2\\
			
			Wang \cite{wang2014weakly} &48.8 &41.0 &23.6 &12.1 &11.1 &42.7 &40.9 &35.5 &11.1 &36.6 &18.4 &35.3 &34.8 &51.3 &17.2 &17.4 &26.8 &32.8 &35.1 &45.6 &30.9\\
			
			Li \cite{li2016weakly} & 54.5 &47.4 &41.3 &20.8 &17.7 &51.9 &63.5 &46.1 &21.8 &57.1 &22.1 &34.4 &50.5 &61.8 &16.2 &29.9 &40.7 &15.9 &55.3 &40.2 &39.5\\
			
			Bilen \cite{bilen2016weakly} & 46.4 &58.3 &35.5 &25.9 &14.0 &66.7 &53.0 &39.2 &8.9 &41.8 &26.6 &38.6 &44.7 &59.0 &10.8 &17.3 &40.7 &49.6 &56.9 &50.8 &39.3\\
			
			Teh \cite{teh2016attention} & 48.8 & 45.9 & 37.4 & 26.9 & 9.2 & 50.7 & 43.4 & 43.6 & 10.6 & 35.9 & 27.0 & 38.6 & 48.5 & 43.8 & 24.7 & 12.1 & 29.0 & 23.2 & 48.8 & 41.9 & 34.5 \\
			
			Tang \cite{tang2017multiple}  & 58.0  & 62.4  & 31.1  & 19.4  & 13.0  & 65.1  & 62.2  & 28.4  & 24.8   & 44.7  & 30.6   & 25.3  & 37.8  & 65.5  & 15.7  & 24.1  & 41.7  & 46.9  & 64.3  & 62.6  & 41.2 \\
			
			Jie \cite{jie2017deep}  & 52.2  & 47.1  & 35.0  & 26.7  & 15.4  & 61.3  & 66.0  & 54.3  & 3.0   & 53.6  & 24.7   & 43.6  & 48.4  & 65.8  & 6.6  & 18.8  & 51.9  & 43.6  & 53.6  & 62.4  & 41.7 \\
			
			Diba \cite{diba2016weakly}  & 49.5	&60.6	&38.6	&29.2&	16.2	&70.8	&56.9	&42.5	&10.9	&44.1	&29.9	&42.2	&47.9	&64.1	&13.8	&23.5	&45.9	&54.1	&60.8	&54.5	&{$\color{blue}\bm{42.8}$} \\
			Lai \cite{lai2017saliency} & 48.4 & 61.5 & 33.3 & 30.0 & 15.3 & 72.4 & 62.4 & 59.1 & 10.9 & 42.3 & 34.3 & 53.1 & 48.4 & 65.0 & 20.5 & 16.6 & 40.6 & 46.5 & 54.6 & 55.1 & {$\color{red}\bm{43.5}$} \\
			\midrule
			%$P_I$  & 54.8  & 64.3 & 37.5  & 28.7  & 13.9  & 63.7  & 62.4  & 47.3  & 16.7  & 45.5  & 29.6  & 26.6  & 41.4  & 63.1  & 10.1  & 23.0 & 42.5  & 50.5  & 63.3  & 57.9  & 42.2 \\
			TS\textsuperscript{2}C  & 59.3  & 57.5  & 43.7  & 27.3  & 13.5  & 63.9  & 61.7  & 59.9  & 24.1  & 46.9  & 36.7  & 45.6  & 39.9  & 62.6  & 10.3  & 23.6  & 41.7  & 52.4  & 58.7  & 56.6  & {$\color{darkgreen}\bm{44.3}$} \\
			\midrule \midrule
			\multicolumn{22}{l}{Comparisons on VOC 2012:}  \\
			Kantorov \cite{kantorov2016contextlocnet} & 64.0 & 54.9 & 36.4 & 8.1 & 12.6 & 53.1 & 40.5 & 28.4 & 6.6 & 35.3 & 34.4 & 49.1 & 42.6 & 62.4 & 19.8 & 15.2 & 27.0 & 33.1 & 33.0 & 50.0 & 35.3 \\		
			Tang \cite{tang2017multiple}  & -  & -  & -  & -  & - & -  & -  & -  & -   & -  & -  & -  & -  & - & -  & -  & - & -  & -  & -  & {$\color{blue}\bm{37.9}$}\\		
			Jie \cite{jie2017deep}  & 60.8 & 54.2 & 34.1 & 14.9 & 13.1 & 54.3 & 53.4 & 58.6 & 3.7 & 53.1 & 8.3 & 43.4 & 49.8 & 69.2 & 4.1 & 17.5 & 43.8 & 25.6 & 55.0 & 50.1 & {$\color{red}\bm{38.3}$} \\	
			\midrule
			TS\textsuperscript{2}C & 67.4  & 57.0  & 37.7  & 23.7  & 15.2  & 56.9  & 49.1  & 64.8  & 15.1  & 39.4  & 19.3  & 48.4  & 44.5  & 67.2  & 2.1  & 23.3  & 35.1  & 40.2  & 46.6  & 45.8  & {$\color{darkgreen}\bm{40.0}$} \\
			\bottomrule
	\end{tabular}}
%	\vspace{-3mm}
\end{table*}

\begin{table}[t]\setlength{\tabcolsep}{2.4pt}
	\centering
	\caption{Comparison of detection AP (\%) by training FRCNN detectors.}
	\begin{tabular}{lcc}
		\toprule
		Method & VOC 2007 & VOC 2012 \\
		\midrule
		TS\textsuperscript{2}C + FRCNN & 48.0 & 44.4 \\
		OICR-Ens. + FRCNN~\cite{tang2017multiple} & 47.0 & 42.5 \\
		\bottomrule
	\end{tabular}
	\label{tab:frcnn}
\end{table}
\begin{figure*}[htb]
	\centering
	\includegraphics[width=0.8\textwidth]{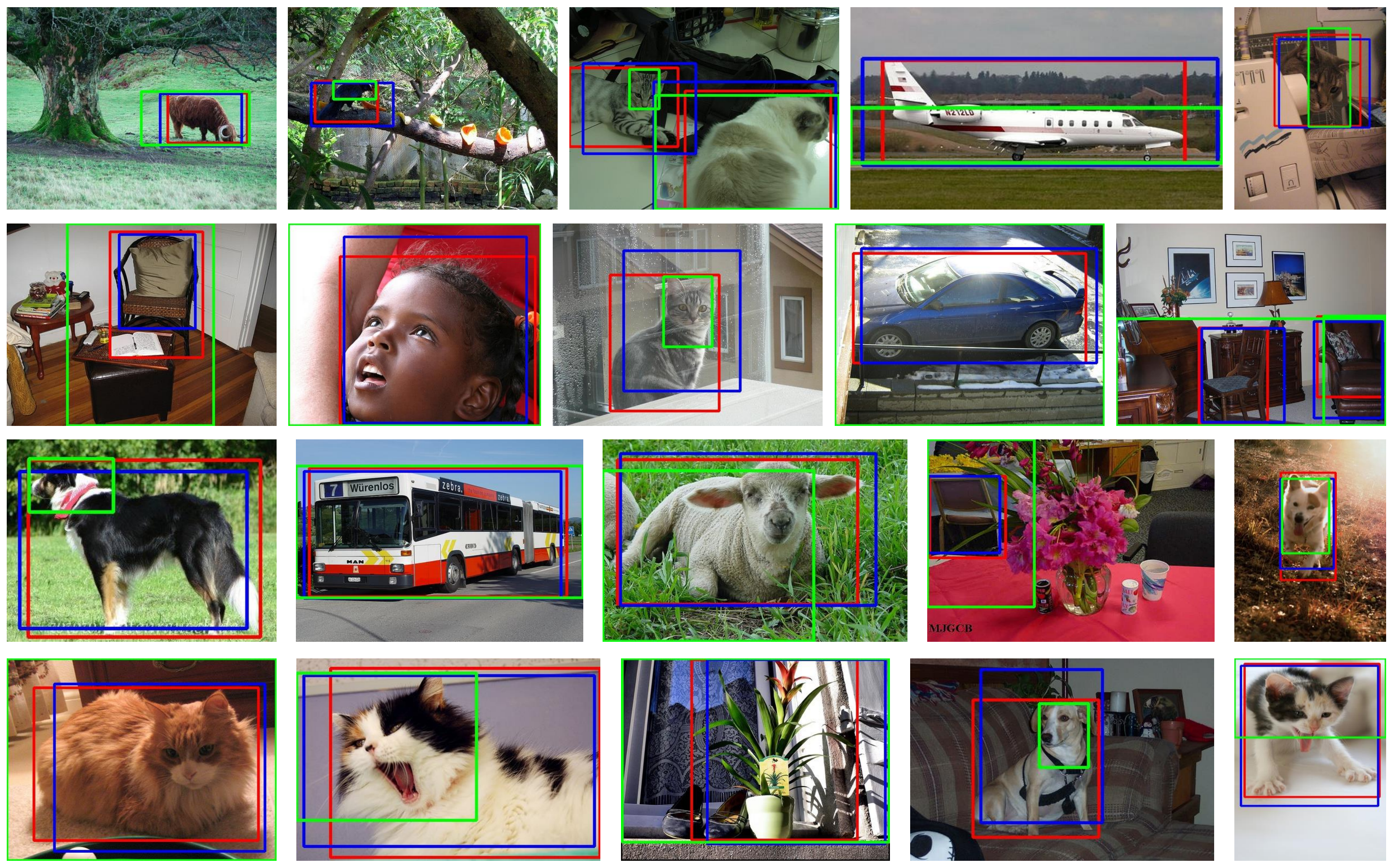}
	\caption{Examples of our object detection results on VOC 2007 test set. Ground-truth annotations, predictions of OICR and ours are indicated by red, green and blue bounding boxes respectively. Best viewed in color.}
	\label{fig:samples}
%	\vspace{-3mm}
\end{figure*}

\begin{table*}[htb]\setlength{\tabcolsep}{2.4pt}
	\centering
	\caption{Comparison of correct localization (CorLoc) (\%) on PASCAL VOC.}
	\label{tab:loc}
	\resizebox{1\textwidth}{!}{
		\begin{tabular}{lcccccccccccccccccccc|c}
			\toprule
			Method & \rotatebox{90}{plane} &   \rotatebox{90}{bike} &   \rotatebox{90}{bird} &   \rotatebox{90}{boat} &   \rotatebox{90}{bottle} &   \rotatebox{90}{bus} &   \rotatebox{90}{car} &   \rotatebox{90}{cat} &   \rotatebox{90}{chair} &   \rotatebox{90}{cow} &   \rotatebox{90}{table} &   \rotatebox{90}{dog} &   \rotatebox{90}{horse} &   \rotatebox{90}{motor} &   \rotatebox{90}{person} &   \rotatebox{90}{plant} &   \rotatebox{90}{sheep} &   \rotatebox{90}{sofa} &   \rotatebox{90}{train} &   \rotatebox{90}{tv} &  \rotatebox{90}{mean}  \\
			\midrule
			\multicolumn{22}{l}{Comparisons on VOC 2007:}  \\
			Bilen \cite{bilen2015weakly} & 66.4 &59.3 &42.7 &20.4 &21.3& 63.4& 74.3& 59.6& 21.1 &58.2 &14.0 &38.5 &49.5 &60.0 &19.8& 39.2 &41.7 &30.1 &50.2 &44.1 &43.7\\
			
			Cinbis \cite{cinbis2017weakly} & 65.3 &55.0& 52.4& 48.3 &18.2 &66.4 &77.8 &35.6& 26.5 &67.0 &46.9 &48.4 &70.5 &69.1& 35.2 &35.2 &69.6 &43.4 &64.6 &43.7 &52.0\\
			
			Wang. \cite{wang2014weakly} &80.1& 63.9& 51.5& 14.9& 21.0& 55.7& 74.2 &43.5 &26.2 &53.4 &16.3 &56.7 &58.3& 69.5 &14.1 &38.3 &58.8 &47.2 &49.1 &60.9 &48.5\\
			
			Li \cite{li2016weakly} & 78.2 &67.1& 61.8 &38.1& 36.1& 61.8 &78.8 &55.2& 28.5 &68.8 &18.5& 49.2 &64.1 &73.5& 21.4 &47.4& 64.6 &22.3 &60.9 &52.3 &52.4\\
			
			Bilen \cite{bilen2016weakly} & 65.1 &63.4& 59.7 &45.9 &38.5 &69.4 &77.0& 50.7 &30.1 &68.8 &34.0 &37.3& 61.0 &82.9& 25.1 &42.9& 79.2 &59.4& 68.2 &64.1 &56.1\\
			
			Jie \cite{jie2017deep}  & 72.7  & 55.3  & 53.0  & 27.8  & 35.2  & 68.6  & 81.9  & 60.7  & 11.6   & 71.6  & 29.7   & 54.3  & 64.3  & 88.2  & 22.2  & 53.7  & 72.2  & 52.6  & 68.9  & 75.5  & 56.1 \\
			
			Diba \cite{diba2016weakly}  & 83.9&	72.8	&64.5&	44.1&	40.1	&65.7	&82.5	&58.9	&33.7	&72.5&	25.6	&53.7&	67.4	&77.4	&26.8	&49.1	&68.1	&27.9	&64.5	&55.7&	56.7 \\
			
			Tang \cite{tang2017multiple}  & 81.7  & 80.4  & 48.7  & 49.5  & 32.8  & 81.7  & 85.4  & 40.1  & 40.6   & 79.5  & 35.7   & 33.7  & 60.5  & 88.8  & 21.8  & 57.9  & 76.3  & 59.9  & 75.3  & 81.4  & 60.6 \\
			Lai \cite{lai2017saliency} & 71.0 & 76.5 & 54.9 & 49.7 & 54.1 & 78.0 & 87.4 & 68.8 & 32.4 & 75.2 & 29.5 & 58.0 & 67.3 & 84.5 & 41.5 & 49.0 & 78.1 & 60.3 & 62.8 & 78.9 & {$\color{red}\bm{62.9}$} \\
			Teh \cite{teh2016attention} & 84.0 & 64.6 & 70.0 & 62.4 & 25.8 & 80.7 & 73.9 & 71.5 & 35.7 & 81.6 & 46.5 & 71.3 & 79.1 & 78.8 & 56.7 & 34.3 & 69.8 & 56.7 & 77.0 & 72.7 & {$\color{darkgreen}\bm{64.6}$} \\
			\midrule
			TS\textsuperscript{2}C  & 84.2  & 74.1  & 61.3  & 52.1  & 32.1  & 76.7  & 82.9  & 66.6  & 42.3  & 70.6  & 39.5  & 57.0  & 61.2  & 88.4  & 9.3  & 54.6  & 72.2  & 60.0  & 65.0  & 70.3  & {$\color{blue}\bm{61.0}$} \\
			\midrule \midrule
			\multicolumn{22}{l}{Comparisons on VOC 2012:}  \\
			Kantorov \cite{kantorov2016contextlocnet} & 78.3 & 70.8 & 52.5 & 34.7 & 36.6 & 80.0 & 58.7 & 38.6 & 27.7 & 71.2 & 32.3 & 48.7 & 76.2 & 77.4 & 16.0 & 48.4 & 69.9 & 47.5 & 66.9 & 62.9 & 54.8 \\
			Jie \cite{jie2017deep}  & 82.4 & 68.1& 54.5& 38.9& 35.9& 84.7& 73.1& 64.8& 17.1& 78.3& 22.5& 57.0& 70.8& 86.6& 18.7& 49.7& 80.7& 45.3& 70.1& 77.3& {$\color{blue}\bm{58.8}$} \\
			Tang \cite{tang2017multiple}  & -  & -  & -  & -  & - & -  & -  & -  & -   & -  & -  & -  & -  & - & -  & -  & - & -  & -  & -  & {$\color{red}\bm{62.1}$}\\
			\midrule
			TS\textsuperscript{2}C & 79.1 & 83.9 &64.6 &50.6&37.8&87.4&74.0&74.1&40.4&80.6&	42.6&53.6&66.5&88.8&18.8&54.9&80.4&60.4&70.7&79.3&{$\color{darkgreen}\bm{64.4}$} \\
			\bottomrule
	\end{tabular}}
\end{table*}
\subsection{Comparison with Other State-of-the-arts}
We compare our approach with both two-step~\cite{wang2014weakly,cinbis2017weakly,li2016weakly,jie2017deep} and end-to-end~\cite{bilen2016weakly,kantorov2016contextlocnet,tang2017multiple,diba2016weakly,lai2017saliency,teh2016attention} approaches. Top-3 results are indicated by ${\color{darkgreen}\bm{green}}$, ${\color{red}\bm{red}}$ and ${\color{blue}\bm{blue}}$ colors. Table~\ref{tab:det} shows the comparison in terms of AP on the VOC 2007. It can be observed that the proposed TS\textsuperscript{2}C is effective and outperforms all the other approaches. In particular, we adopt OICR proposed by Tang \emph{et al.}~\cite{tang2017multiple} as the detection backbone in the proposed framework. Our approach outperforms OICR by 3.1\%. The gains are mainly from using both purity and completeness metrics for filtering noisy object candidates. We also show the comparison between our approach and other state-of-the-arts on PASCAL VOC 2012 in terms of AP. Our result\footnote{http://host.robots.ox.ac.uk:8080/anonymous/GDNUDG.html} outperforms the baseline (\ie Tang \etal \cite{tang2017multiple}) and the state-of-the-art approach (\ie Jie \etal \cite{jie2017deep}) by 2.1\% and 1.7\%, respectively.  

Following~\cite{tang2017multiple}, we also train a FRCNN~\cite{girshick2015fast} detector using top-scoring proposals produced by TS\textsuperscript{2}C as pseudo ground-truth bounding boxes. As shown in Table~\ref{tab:frcnn}, the performance can be further enhanced to 48.0\% and 44.4\%\footnote{http://host.robots.ox.ac.uk:8080/anonymous/ECKWR7.html} on VOC 2007 and 2012, respectively. Our results from a single model are much better than those of~\cite{tang2017multiple} obtained by models (\eg VGG16 and VGG-M) fusion.
In addition, we conduct additional experiments using CorLoc as the evaluation metric. Table~\ref{tab:loc} shows the comparison on the VOC 2007 and 2012. Our approach achieves 61.0\% and 64.4\% in terms of CorLoc score, which are competitive compared with the state-of-the-arts. We visualize some successful detection results (blue boxes) on VOC 2007, as shown in Figure~\ref{fig:samples}. Results from OICR (green boxes) and ground truth (red boxes) are employed for comparison. It can be seen that our approach effectively reduces false positives including partial objects.

%\begin{figure*} 
%	\centering
%	\includegraphics[width=0.6\textwidth]{fig2/recall.pdf}
%	\caption{Comparison of recall scores (\%) between the proposed TS\textsuperscript{2}C ($P_I-P_S$) and the purity strategy ($P_I$).}
%	\label{fig:recall}
%\end{figure*}
\subsection{Ablation Experiments}

We conduct extensive ablation analyses of the proposed TS\textsuperscript{2}C, including the influence of the enlarged scale for obtaining surrounding context and the proposed tightness criteria (\ie purity and completeness). All experiments are based on VOC 2007 benchmark.

%\begin{figure}
%	\hfill\begin{minipage}{0.5\textwidth}\centering
%		\includegraphics[width=1\textwidth]{fig2/recall.pdf}
%		\caption{Comparison of recall scores (\%) between the proposed TS\textsuperscript{2}C ($P_I-P_S$) and the purity strategy ($P_I$).}
%	\end{minipage}
%\label{fig:recall}
%\end{figure}

\begin{table*}[htb]\setlength{\tabcolsep}{3.1pt}
	\centering
	\caption{Ablation study on PASCAL VOC 2007.}
	\label{tab:scale}
	\resizebox{1\textwidth}{!}{
		\begin{tabular}{lccccccccccccccccccccc}
			\toprule
			Method & \rotatebox{90}{plane} &   \rotatebox{90}{bike} &   \rotatebox{90}{bird} &   \rotatebox{90}{boat} &   \rotatebox{90}{bottle} &   \rotatebox{90}{bus} &   \rotatebox{90}{car} &   \rotatebox{90}{cat} &   \rotatebox{90}{chair} &   \rotatebox{90}{cow} &   \rotatebox{90}{table} &   \rotatebox{90}{dog} &   \rotatebox{90}{horse} &   \rotatebox{90}{motor} &   \rotatebox{90}{person} &   \rotatebox{90}{plant} &   \rotatebox{90}{sheep} &   \rotatebox{90}{sofa} &   \rotatebox{90}{train} &   \rotatebox{90}{tv} &  \rotatebox{90}{mAP}  \\
			\midrule
			\multicolumn{22}{l}{$P_I$ \emph{vs.} $P_I - P_S$:}  \\ 
			$P_I$  & 54.8  & 64.3 & 37.5  & 28.7  & 13.9  & 63.7  & 62.4  & 47.3  & 16.7  & 45.5  & 29.6  & 26.6  & 41.4  & 63.1  & 10.1  & 23.0 & 42.5  & 50.5  & 63.3  & 57.9  & 42.2 \\
			$P_I - P_S$ & 59.3  & 57.5  & 43.7  & 27.3  & 13.5  & 63.9  & 61.7  & 59.9  & 24.1  & 46.9  & 36.7  & 45.6  & 39.9  & 62.6  & 10.3  & 23.6  & 41.7  & 52.4  & 58.7  & 56.6  & $\bm{44.3}$ \\
			\midrule
			\multicolumn{22}{l}{Enlarged scales:}  \\ 
			baseline & 58.0  & 62.4  & 31.1  & 19.4  & 13.0  & 65.1  & 62.2  & 28.4  & 24.8   & 44.7  & 30.6   & 25.3  & 37.8  & 65.5  & 15.7  & 24.1  & 41.7  & 46.9  & 64.3  & 62.6  & 41.2 \\
			scale (1.4)  & 61.3  & 58.1  & 44.7  & 26.2  & 10.1  & 65.0 & 60.5 & 37.2  & 28.3  & 49.8  & 40.9  & 24.2  & 38.9  & 62.1  & 9.4  & 23.9  & 41.7  & 51.0  & 60.8  & 58.8  & 42.6 \\
			scale (1.3)  & 61.2  & 60.2  & 39.7  & 29.0  & 9.8   & 65.2  & 59.5  & 53.3  & 24.5  & 48.3  & 41.0  & 33.9  & 40.4  & 61.4  & 12.2  & 22.5  & 42.1  & 52.5  & 59.4  & 60.9  & 43.8 \\	
			scale (1.2)  & 59.3  & 57.5  & 43.7  & 27.3  & 13.5  & 63.9  & 61.7  & 59.9  & 24.1  & 46.9  & 36.7  & 45.6  & 39.9  & 62.6  & 10.3  & 23.6  & 41.7  & 52.4  & 58.7  & 56.6  & $\bm{44.3}$ \\
			scale (1.1)  & 59.6  & 58.1  & 41.3  & 29.1  & 13.3  & 64.0  & 60.6  & 52.9  & 25.7  & 49.9  & 45.6  & 29.2  & 40.4  & 61.4  & 11.6  & 22.9  & 40.8  & 48.3  & 60.3  & 60.7  & 43.8 \\
			\midrule
			\multicolumn{22}{l}{Conditional average strategy:}  \\ 
			top 30\% & 60.8  & 58.7 & 39.7  & 33.2  & 11.2  & 64.3  & 60.5  & 52.6  & 24.8  & 48.1  & 37.2  & 25.6  & 45.5  & 63.7  & 11.4  & 23.8 & 40.9  & 49.1  & 58.4  & 59.9  & 43.5 \\
			top 50\% & 59.3  & 57.5  & 43.7  & 27.3  & 13.5  & 63.9  & 61.7  & 59.9  & 24.1  & 46.9  & 36.7  & 45.6  & 39.9  & 62.6  & 10.3  & 23.6  & 41.7  & 52.4  & 58.7  & 56.6  & $\bm{44.3}$ \\
			top 70\% & 60.9  & 61.5  & 41.8  & 31.8  & 12.8  & 64.8  & 60.3  & 46.5  & 22.8  & 49.7  & 38.7  & 26.3  & 50.2  & 63.2  & 12.7  & 22.4  & 41.6  & 49.4  & 60.0  & 60.3  & {43.9} \\    
			all pixels & 60.8  & 60.3  & 38.2  & 31.2  & 11.3  & 63.6  & 60.1  & 55.6  & 20.9  & 51.9  & 40.0  & 33.4  & 41.2  & 64.6  & 11.1  & 23.2  & 43.0  & 47.7  & 59.6  & 59.3  & {43.8} \\
			
			\bottomrule
	\end{tabular}}
\end{table*}

% \begin{figure}
% %	\vspace{-5mm}
% 	\centering
% 	\includegraphics[width=0.45\textwidth]{fig2/recall.pdf}
% 	\caption{Comparison of recall scores (\%) between the proposed TS\textsuperscript{2}C ($P_I-P_S$) and the purity strategy ($P_I$).}
% 	\label{fig:recall}
% %	\vspace{-6mm}
% \end{figure}

% \begin{wrapfigure} {r} {0.45\textwidth}
% %	\vspace{-5mm}
% 	\includegraphics[width=0.45\textwidth]{fig2/recall.pdf}
% 	\caption{Comparison of recall scores (\%) between the proposed TS\textsuperscript{2}C ($P_I-P_S$) and the purity strategy ($P_I$).}
% 	\label{fig:recall}
% %	\vspace{-6mm}
% \end{wrapfigure}

\subsubsection{Purity and Completeness}
\begin{figure}[t]
  \centering
    \begin{minipage}{0.45\linewidth}
      \includegraphics[width=1\linewidth]{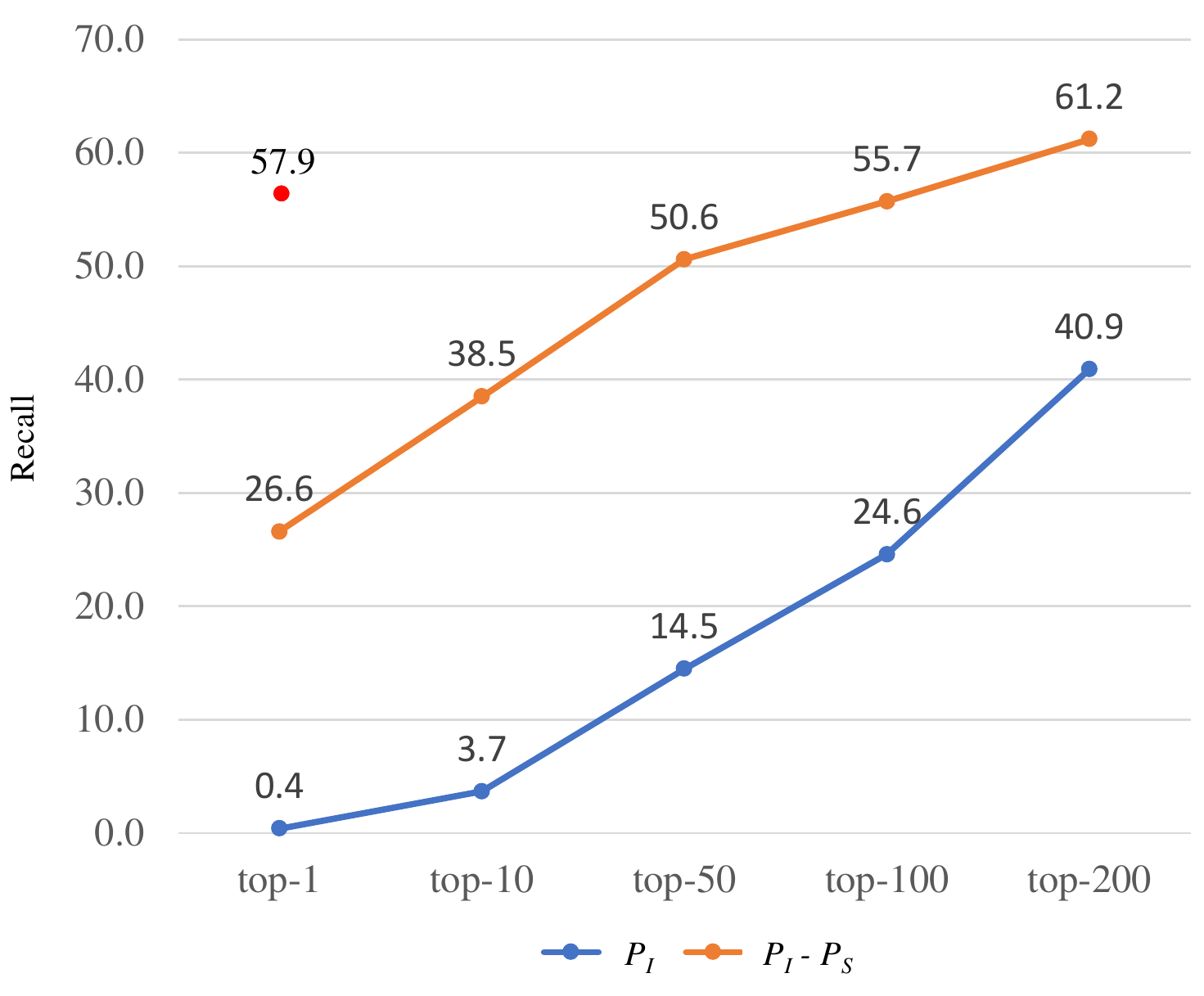}
    \end{minipage}\hfill
    \begin{minipage}{0.45\linewidth}
      \caption{Comparison of recall scores (\%) between the proposed TS\textsuperscript{2}C ($P_I-P_S$) and the purity strategy ($P_I$).\label{fig:recall}}
    \end{minipage}
\end{figure}

One of our main contributions is the proposed criteria of purity and completeness for measuring the tightness of object candidates based on the semantic segmentation confidence maps. To validate the effectiveness of our approach (\ie $P_I-P_S$), we test the other popular setting where only the purity (\eg $P_I$) is taken into account. Specifically, we firstly leverage the two metrics to rank object candidates for annotated class(es). For example, if the image is annotated with two labels, we will produce two rankings according to segmentation confidence maps of the two classes, which are then employed for evaluating recall scores. As shown in Figure~\ref{fig:recall}, we vary the top number of object candidates based on the rankings from two metrics. Since our evaluation method only takes one object candidate for each annotated category in the top-1 case, the upper bound of the recall is 57.9\% due to the existence of multi-instance images. Despite the apparent simplicity, the recall scores of our proposed $P_I-P_S$ significantly outperform those of $P_I$ under different settings according to the top number, which demonstrates that the completeness metric is effective for reducing noisy object candidates. More visualizations of rank 1 boxes produced by $P_I-P_S$ and $P_I$ are shown in Figure~\ref{fig:metric}. We can observe that our approach can successfully discover the tight ones from thousands of candidates. To further validate the effectiveness of the proposed TS\textsuperscript{2}C, we also conduct experiments using purity \ie $P_I$ for ranking object candidates as adopted in~\cite{diba2016weakly} for proposal selection, which results in 42.2\% in mAP. By simultaneously taking purity and completeness into account, \ie  $P_I-P_S$, the result surpasses the baseline by 2.1\% as shown in Table~\ref{tab:scale}.

\subsubsection{Influence of Enlarged Scale}
To evaluate the completeness of object candidates, we need to enlarge the original box with a specific ratio. As shown in Table~\ref{tab:scale}, we examine four ratios (\ie from 1.1 to 1.4) for obtaining the surrounding context of object candidates, which are then employed to calculate objectness scores with the proposed TS\textsuperscript{2}C. We can observe that all the models trained with the proposed TS\textsuperscript{2}C can outperform the baseline by more than 1.4\%. In particular, the best result is achieved by adopting the ratio of 1.2. By continually enlarging the ratio, the performance will be decreased. The reason may be that some training images include multiple instances with the same semantics, and the completeness score of each object candidate will be influenced by adjacent instances in the case of using larger ratios. 

\subsubsection{Influence of Conditional Averaging Strategy}
As shown in Table~\ref{tab:scale}, we also examine the threshold of conditional average strategy. The best result is achieved by employ the first 50\% largest pixels to calculate the objectness score of surrounding region.

\begin{figure*}[t]
	\centering
	\includegraphics[width=0.88\textwidth]{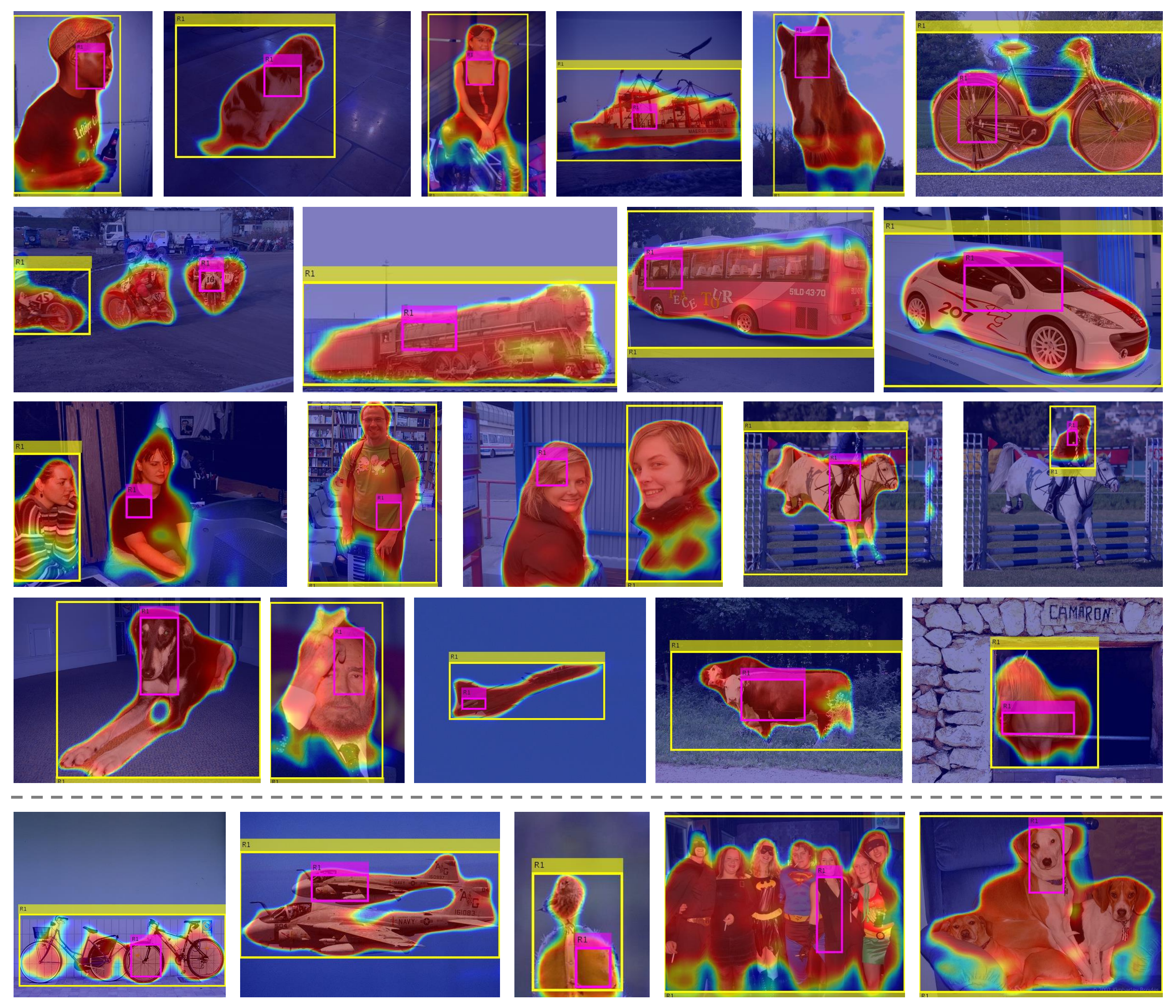}
	\caption{Rank 1 object candidates inferred by the proposed TS\textsuperscript{2}C (yellow boxes) and the strategy only using purity metric for ranking (magenta boxes). Some failure cases are given in the last row. Best viewed in color.}
	\label{fig:metric}
\end{figure*}

\subsubsection{Discussion}
Some failure cases are shown in the last row of Figure~\ref{fig:metric}. These samples share some similar characteristics: low-quality segmentation predictions or many semantically identical instances are linked together. For instance (the middle image of the last row), the semantic segmentation branch makes a false prediction for the object under the \emph{bird}, leading to incorrect inference of our approach. It is believed that such a case can be well addressed with the development of weakly supervised semantic segmentation techniques. For other failure samples, although the segmentation branch can provide high quality confidence maps, the overlap between objects results in false prediction of our TS\textsuperscript{2}C. In this case, we may need to develop effective instance-level semantic segmentation approaches in a weakly supervised manner. 

However, the limitation of our TS\textsuperscript{2}C to deal with overlapping objects with the same semantics does not affect its good performance on WSOD. We do not employ the top-1 proposal according to the objectness score as the object candidate, but build a candidate pool by selecting the top two hundred proposals. In this case, these tight boxes may still be recalled even without the largest tightness score. The effectiveness of our TS\textsuperscript{2}C can be well proved by the performance gains on VOC 2007 and 2012 compared with~\cite{tang2017multiple}.

%-------------------------------------------------------------------------
\section{Conclusion and Future Work}
In this work, we proposed a simple approach, \ie TS\textsuperscript{2}C, for mining tight boxes by exploiting surrounding segmentation context. The TS\textsuperscript{2}C is effective for suppressing low quality object candidates and promoting high quality ones tightly covering the target object. Based on the segmentation confidence map, TS\textsuperscript{2}C introduces two simple criteria, \ie purity and completeness, to evaluate objectness scores of object candidates. Despite apparent simplicity, the proposed TS\textsuperscript{2}C can effectively filter thousands of noisy candidates and be easily embedded into any end-to-end weakly supervised framework for performance improvement. In the future, we plan to design more effective metrics for mining tight boxes by further boosting our current approach.

%\vspace{3mm}
\noindent
\textbf{Acknowledgements} This work is in part supported by IBM-ILLINOIS Center for Cognitive Computing Systems Research (C3SR) - a research collaboration as part of the IBM AI Horizons Network, NUS IDS R-263-000-C67-646, ECRA R-263-000-C87-133, MOE Tier-II R-263-000-D17-112 and the Intelligence Advanced Research Projects Activity (IARPA) via Department of Interior/ Interior Business Center (DOI/IBC) contract number D17PC00341. The U.S. Government is authorized to reproduce and distribute reprints for Governmental purposes notwithstanding any copyright annotation thereon.  Disclaimer: The views and conclusions contained herein are those of the authors and should not be interpreted as necessarily representing the official policies or endorsements, either expressed or implied, of IARPA, DOI/IBC, or the U.S. Government.

\clearpage
\footnotesize{
\bibliographystyle{splncs}
\bibliography{egbib}
}
\end{document}